\documentclass{article}
\usepackage[a4paper, total={6in, 9.5in}]{geometry}
\usepackage[dvipsnames]{xcolor}
\usepackage{mathtools} \usepackage{multirow}
\usepackage{booktabs}
\usepackage{colortbl}
\usepackage{subfigure}
\usepackage[export]{adjustbox}
\usepackage{afterpage}

\usepackage{wrapfig}

\usepackage{caption}

\usepackage[utf8]{inputenc} 
\usepackage[T1]{fontenc}    
\usepackage{hyperref}       
\usepackage{url}            
\usepackage{booktabs}       
\usepackage{amsfonts}       
\usepackage{nicefrac}       
\usepackage{microtype}      
\usepackage{xcolor}         

\usepackage{graphbox,graphicx}

\def\eg{\emph{e.g.~}}

\def\ie{\emph{i.e.~}}
\def\aka{\emph{a.k.a.~}}

\usepackage{tikz}

\title{How good are deep models in understanding\\ the generated images?}

\author{
Ali Borji \\
Quintic AI, San Francisco, CA \\ 
\texttt{aliborji@gmail.com} 
}

\begin{document}

\maketitle

\begin{abstract}
 My goal in this paper is twofold: to study how well deep models can understand the images generated by DALL-E 2 and Midjourney, and to quantitatively evaluate these generative models. Two sets of generated images are collected for object recognition and visual question answering (VQA) tasks. On object recognition, the best model, out of 10 state-of-the-art object recognition models, achieves about 60\% and 80\% top-1 and top-5 accuracy, respectively. These numbers are much lower than the best accuracy on the ImageNet dataset (91\% and 99\%). On VQA, the OFA model scores 77.3\% on answering 241 binary questions across 50 images. This model scores 94.7\% on the binary VQA-v2 dataset. Humans are able to recognize the generated images and answer questions on them easily. We conclude that a) deep models struggle to understand the generated content, and may do better after fine-tuning, and b) there is a large distribution shift between the generated images and the real photographs. The distribution shift appears to be category-dependent. Data is available at: \href{https://drive.google.com/file/d/1n2nCiaXtYJRRF2R73-LNE3zggeU_HeH0/view?usp=sharing}{link}.
    
    
\end{abstract}

\section{Introduction}

Recent deep generative models such as DALL-E 2~\cite{ramesh2022hierarchical} and Midjourney\footnote{\url{https://www.midjourney.com/}} have made a big splash.
They are capable of synthesizing stunning photo-realistic images for a given input text (\aka a prompt), and have inspired many people, in particular the artists. Some researchers have also used these tools to synthesize data for training deep models (\eg~\cite{ge2022dall}). For the most part, the images generated by these systems capture what is included in the input in terms of the objects and their relations. Some studies (\eg~\cite{marcus2022very}\footnote{See \url{https://tinyurl.com/yc7u8juf}, \url{https://tinyurl.com/2p9wku6e}, and \url{https://www.reddit.com/r/dalle2/}.}) have anecdotally and qualitatively inspected these images and have found that they are limited in certain ways. For example, they do not understand the numbers, counting, and negation, have spelling errors, and lack common sense.

On the one hand, deep models such as ResNet~\cite{resnet} are believed to surpass humans in object classification. Here, we test the capability of recent best object classification models on generated images that are easily recognizable by humans. We also investigate the performance of VQA models on answering binary questions on generated images. The outcomes will inform us about the generalization power of deep models.

On the other hand, unlike the significant body of work that has quantitatively evaluated the images generated by GANs~\cite{goodfellow2014generative,borji2019pros,borji2022pros}, little effort has been spent on evaluating DALL-E 2 and Midjourney. The authors of these papers have already used measures such as FID~\cite{heusel2017gans} to quantitatively evaluate their systems. However, research has shown that relying on one score is usually not enough to draw strong conclusions. Here, we take a different approach and argue that if generated images are good, then deep models should be able to recognize them. We feed the synthesized images to the best object recognition models and measure the classification accuracy for each object category. We find that, although the average model performance is poor, models score very high and near perfect over some object categories. This indicates that generative models can capture some categories better than others. Visually inspecting the images from the hard categories, reveals that they are indeed hard to recognize by humans (\eg kites). Admittedly, our results should be taken with a grain of salt, as classification accuracy may favor a generative model that sacrifices sample diversity in favor of generating high fidelity samples. Generative models are expected to generate a diverse set of high-fidelity images and this is what some scores attempt to measure (\eg FID~\cite{heusel2017gans}, or Precision-Recall~\cite{sajjadi2018assessing}).



\begin{figure}[t]
    \centering
    \includegraphics[width=.8\linewidth]{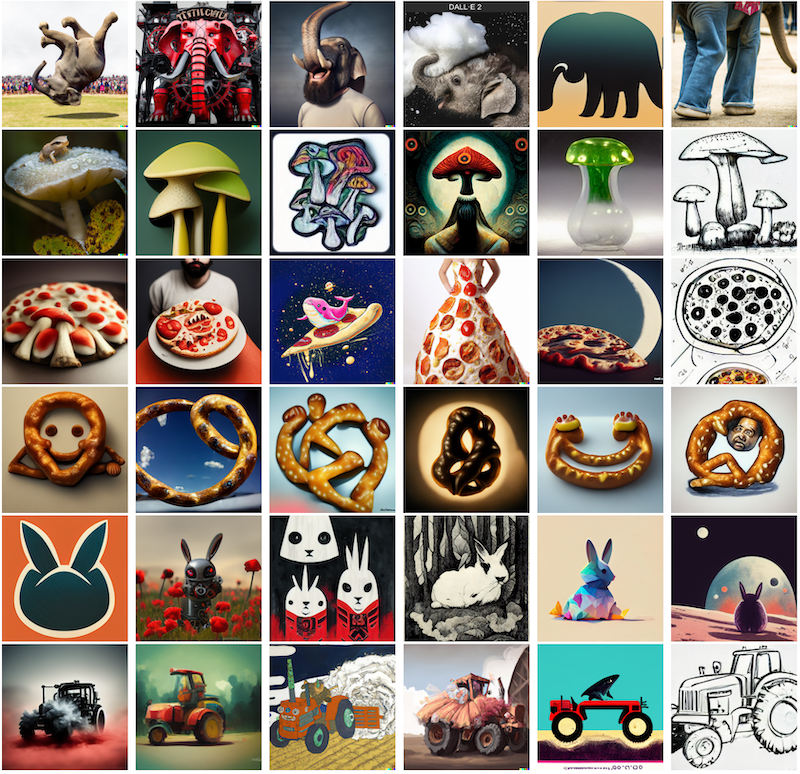}         
    \caption{Sample generated images by DALL-E 2 or Midjourney models. Categories in order are: elephant, mushroom, pizza, pretzel, rabbit, and tractor.}    
    \label{fig:Samples}
\end{figure}

We start by curating a dataset of images generated by DALL-E 2 and Midjourney by crawling images from twitter posts as well as Google search. We did not use images that clearly depict faces of people per DALL-E 2 guidelines\footnote{\url{https://tinyurl.com/r4xeyhps}}. Only images that have good quality and are recognizable by humans are selected. We created two sets of images, one for object recognition and another for visual question answering~\cite{antol2015vqa}. Sample images from these sets are shown in Fig.~\ref{fig:Samples}, and Fig.~\ref{fig:vqa0}, respectively.



\section{Object recognition}

\begin{figure}[t]
    \centering
    \includegraphics[align=t,width=.45\linewidth]{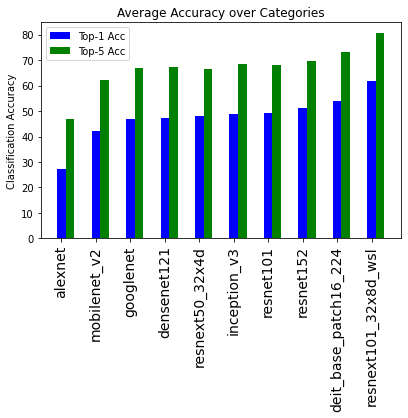}
    \includegraphics[align=t,width=.45\linewidth]{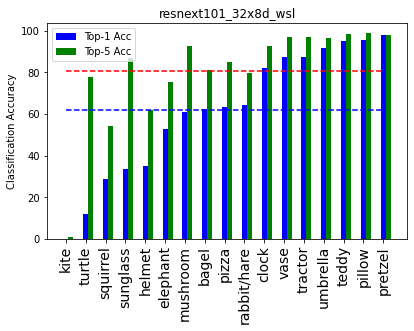}
    \includegraphics[align=t,width=.45\linewidth]{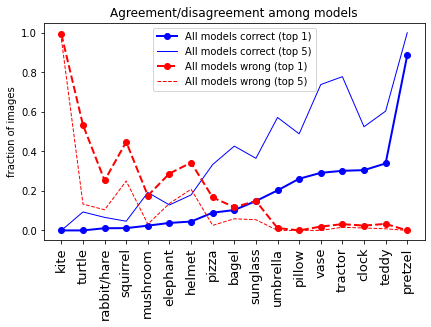}
    \includegraphics[align=t,width=.45\linewidth,height=.33\linewidth]{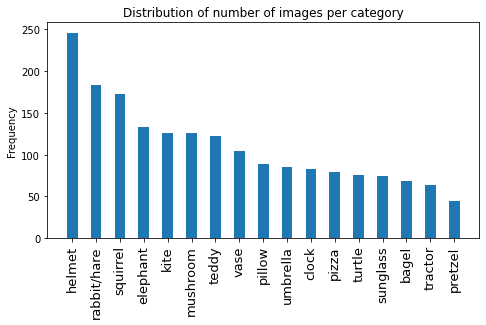}
    \caption{Top left: per model performance of the models averaged over 17 categories. Top right: performance of the best model per category. The dashed lines show the average performance. See Appendix~\ref{appx:model_acc} for performance of individual models. Bottom left: fraction of images over which all models fail or they all succeed. Bottom-right: number of images per category.}
    \label{fig:res}
\end{figure}

We collected 1,862 synthetic images generated by DALL-E 2 and Midjourney across 17 categories (Fig.~\ref{fig:Samples}). The number of images per category is shown in the bottom-right panel of Fig.~\ref{fig:res}. We tested 10 state-of-the-art object recognition models\footnote{Models are available in PyTorch hub: \url{https://pytorch.org/hub/}.}, pre-trained on ImageNet, on these images. These models have been published over the past several years and have been immensely successful over the ImageNet benchmark. They include AlexNet~\cite{krizhevsky2012imagenet}, MobileNetV2~\cite{sandler2018mobilenetv2}, GoogleNet~\cite{szegedy2015going}, DenseNet~\cite{huang2017densely}, ResNext~\cite{xie2017aggregated}, ResNet101~\cite{resnet}, ResNet152~\cite{resnet}, Inception\_V3~\cite{szegedy2016rethinking}, Deit~\cite{touvron2021training}, and ResNext\_WSL~\cite{mahajan2018exploring}. 

Since some of our classes cover multiple ImageNet classes\footnote{\url{https://deeplearning.cms.waikato.ac.nz/user-guide/class-maps/IMAGENET/}}, we had to make some adjustments for computing accuracy. For example, ImageNet has three types of clocks including `digital clock', `wall clock', and `analog clock'. Here, we only have the `clock' class, containing mostly analog clocks. We chose to give the benefit of the doubt to models. A prediction is deemed correct if the ground-truth label is in the set of the words predicted by the model. In the mentioned scenario, if a model predicts `wall clock', then a hit is counted. If the model predicts `wall' or anything else, then the prediction would be considered a mistake. The same is true for the top-5 accuracy computation. For example, if the top five model predictions are `bib', `necklace', `toilet seat', `pick', and `wall clock', then the prediction is counted as a hit. In practice, first all words in the predicted labels are extracted, and then the prediction is counted as a hit if the ground-truth is in this set. In case of ground-truth having two words (\eg `toilet seat'), then it should happen in the set of words exactly as it is. Since `rabbit' and `hare' classes are very similar, we consider both of them to be true predictions. Notice that this way of accuracy measurement gives an overestimation of the model performance, but it is good enough for our purposes here. Even with this overestimation, as we will show, models still perform poorly.

Results are shown in Fig.~\ref{fig:res}. Among the models, \texttt{resnext101\_32x8d\_ws} ranks the best, and significantly better than other models. It achieves around 60\% top-1 and about 80\% top-5 accuracy. This model scores 85.4\% top-1 and 97.6\% top-5 accuracy over the ImageNet-1k validation set (single-crop). The success of this model can be attributed to the fact that it is trained to predict hashtags on billions of social media images in a weakly supervised manner. The best performance on our data is much lower than the best available performance on the ImageNet validation set which are 91\% and 99\% corresponding to top-1 and top-5 accuracy\footnote{\url{https://paperswithcode.com/sota/image-classification-on-imagenet}}. These results suggest that there is a big difference between the distribution of ImageNet images and the distribution of generated images.

According to Fig.~\ref{fig:res}, the top five most difficult categories for the \texttt{resnext101\_32x8d\_ws} model in order are \texttt{kite, turtle, squirrel, sunglass}, and \texttt{helmet}. The performance on these categories is below 40\%. The kite class is often confused with parachute, balloon, and umbrella classes as shown in Fig.~\ref{fig:Errors_kite}. Sample failure cases from the categories along with the predictions are shown in Appendix~\ref{appx:errors}. Models often fail on drawings, unusual objects, or images where the object of interest is not unique.

\begin{figure}[t]
\centering 
 \includegraphics[width=.8\linewidth]{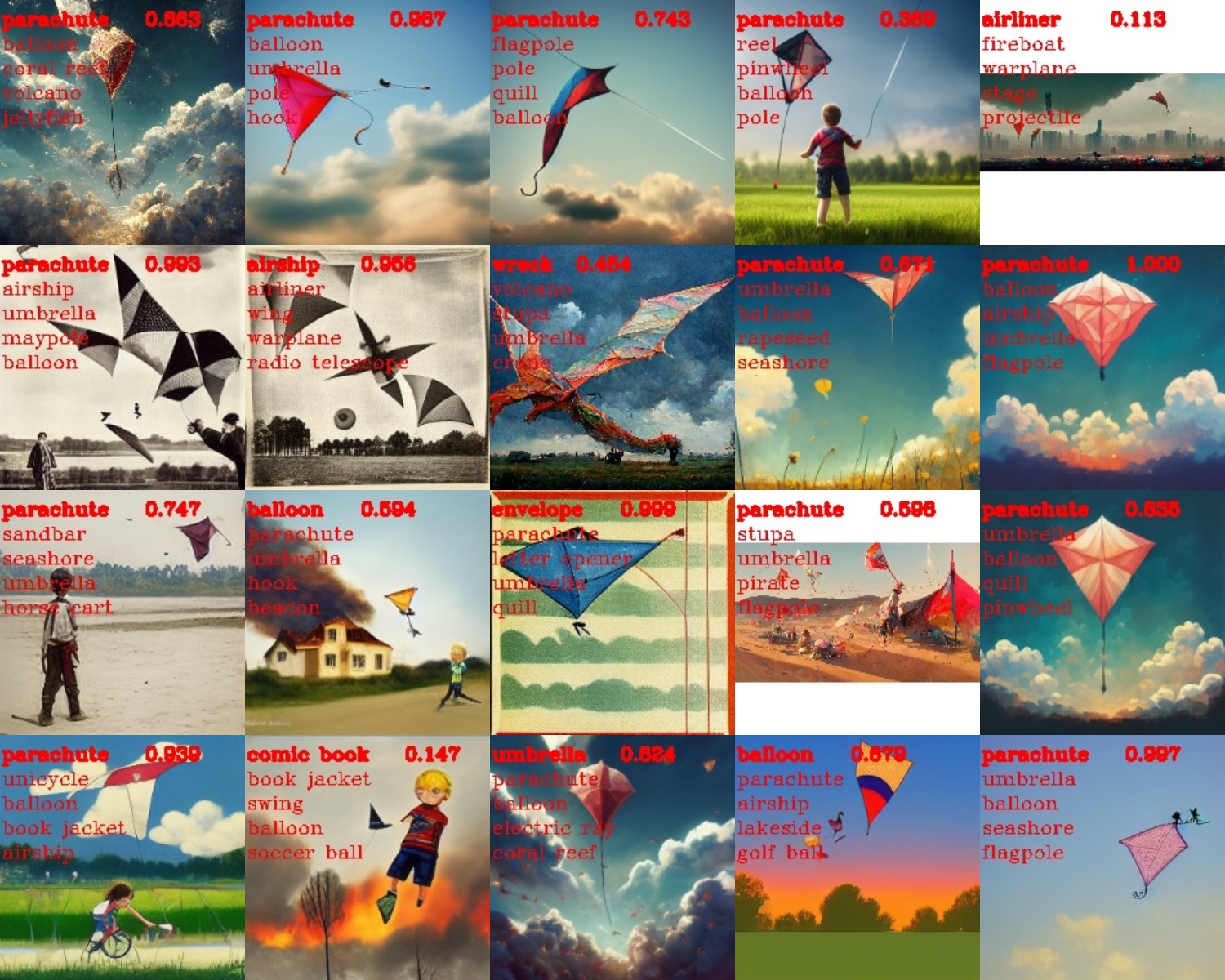}
\caption{Sample images from the kite category over which the resnext101\_32x8d\_wsl model fails. An image is considered an error if the ground-truth is not within the top 5 predictions.} 
\label{fig:Errors_kite}
\end{figure}

 We also computed the fraction of images, per category, over which all models succeed, or they all fail. Results are shown in the bottom left panel of Fig.~\ref{fig:res}. We noticed that for some categories such as kite and turtle models consistently fail, while for some others such as pretzel and tractor they all do very well. When all models succeed, they are correct at best over 90\% of the images (over pretzel category using top-1 acc). These results indicate that models share similar weaknesses and strengths.


\section{Visual question answering}

Here, we test VQA models on free-form and open-ended visual question answering. We only consider binary questions since in principle, any question can be converted to a binary one on an image. Recent VQA models are able to answer binary questions above 95\% accuracy over the VQA-v2 dataset\footnote{https://visualqa.org/}, which is astonishing considering the complexity of the questions.

We collected 50 images and formulated a total of 241 questions on them. There are 4.82 questions per image on average. 132 questions have positive answers and 109 have negative answers. Average number of words per question is 5.12 (\ie question length).

To see how well the state-of-the-art VQA models perform on the generated images, we choose the OFA model~\cite{wang2022ofa} which is currently the leading scorer on the VQA-v2 test-std set\footnote{https://paperswithcode.com/sota/visual-question-answering-on-vqa-v2-test-std}. This model achieves 77.27\% accuracy on generated images. To put this result in perspective, this model scores about 94.7\% on the VQA-v2 test-std set. There are two reasons why OFA performs lower here a) generated images may contain semantic content that is missing in the training set of the VQA-v2 dataset (\eg `the astronaut riding the horse'), and b) our questions might be more challenging than VQA-v2 questions. We suspect the first reason is more viable, which is also in alignment with our earlier observations on visual recognition.
Sample images and questions along with predictions of the OFA model are shown in Fig.~\ref{fig:vqa0}. Additional images are shown in Appendix~\ref{appx:vqa}.


\begin{figure}[t]
\centering 
 \includegraphics[width=\linewidth]{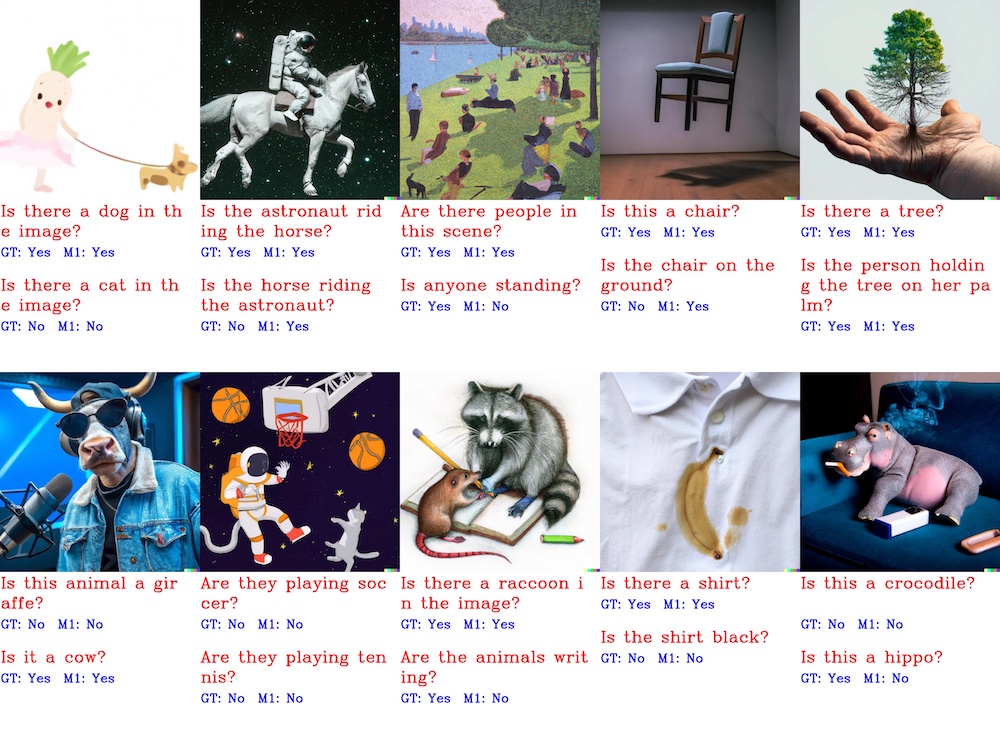}
 \vspace{-20pt}
\caption{Sample images and questions along with the predictions of the OFA model.} 
\label{fig:vqa0}
\vspace{-5pt}
\end{figure}

\section{Discussion and conclusion}
We tested deep models on generated images over two tasks of object recognition and visual question answering. Models perform poorly on these data compared to their performance on real images over which they have been trained on (ImageNet and VQA-v2). We conclude that a) generative models synthesize images for some categories better than other categories, and b) there is a large distribution shift between real images and synthetic images, and this is perhaps why deep models struggle over the latter. We foresee four directions for future research in this area: 
\begin{enumerate}
    \item We did not distinguish between images generated by DALL-E 2 and Midjourney. A quantitative comparison between the two models would be interesting. 
    \item We tested models on a small test set of generated images. Results over a larger set of images and object classes are likely to provide more insights.
    \item It is hard to tell from our results whether low performance on the generated images is due to problems with the images (\eg low fidelity, artifacts, etc), or lack of generalization by the classification models. Visual inspection supports the latter. One way to address this shortcoming is to train and test a deep classifier on generated images. If such a classifier performs well, then it hints towards the high quality of the generated images (and vice-versa).  
    \item Generative models offer a unique opportunity to automatically generate large scale data for training data-hungry deep models. It would be interesting to see how well models trained on synthetic data generated by DALL-E 2 and Midjourney generalize to real-world data. See~\cite{ge2022dall} as an example in the context of object detection.
\end{enumerate}

\bibliographystyle{plain}
\bibliography{refs}

\newpage
\appendix

\newpage
\section{Performance of the individual models}
\label{appx:model_acc}

\begin{figure}[htbp]
    \vspace{20pt}
    \centering
    \includegraphics[width=.45\linewidth]{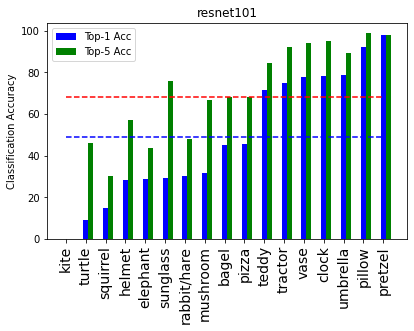} \hspace{15pt}
    \includegraphics[width=.45\linewidth]{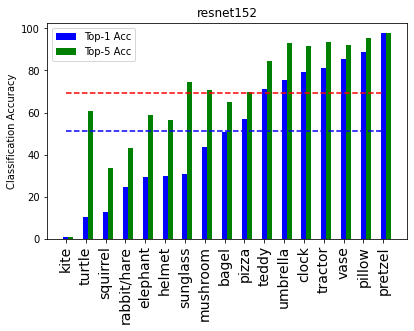} \\
    \includegraphics[width=.45\linewidth]{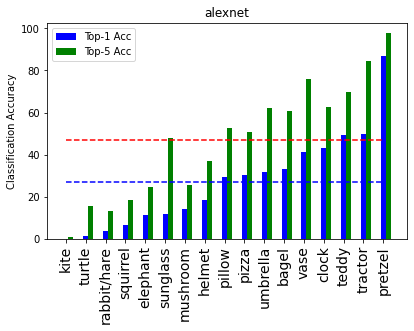} \hspace{15pt}
    \includegraphics[width=.45\linewidth]{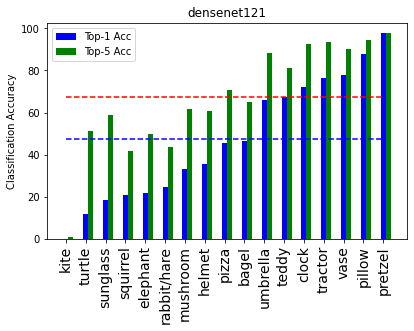} \\
    \includegraphics[width=.45\linewidth]{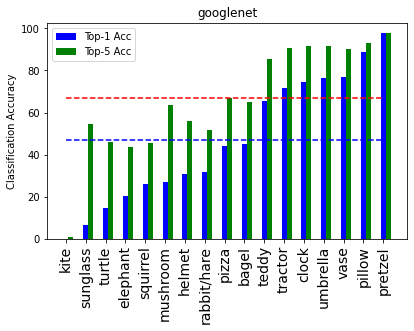} \hspace{15pt}
    \includegraphics[width=.45\linewidth]{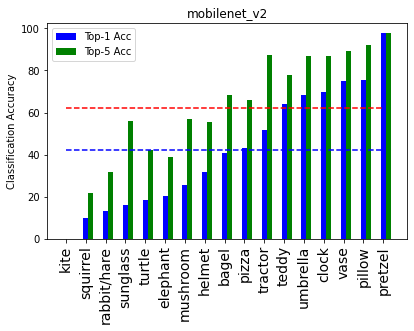} 
    \caption{Performance of individual models on object detection over generated images.}
    \label{fig:appx_D2O_models1}
\end{figure}

\begin{figure}[htbp]
    \centering
    \includegraphics[width=.45\linewidth]{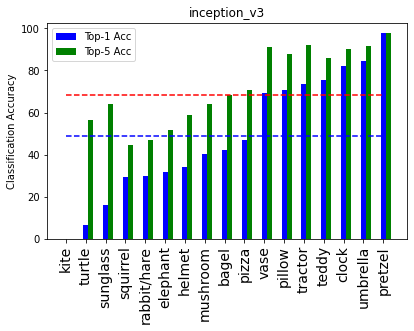} \hspace{15pt}
    \includegraphics[width=.45\linewidth]{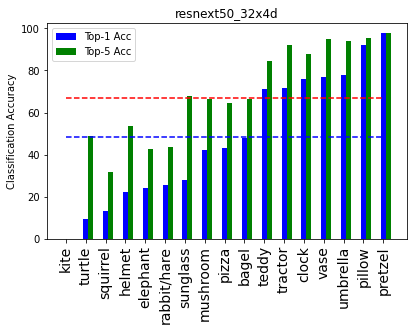} \\
    \includegraphics[width=.45\linewidth]{Figs/Res/9.png} \hspace{15pt}
    \includegraphics[width=.45\linewidth]{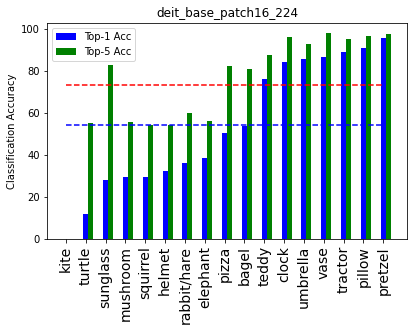} 
    \caption{Performance of individual models on object detection over generated images (cnt'd).}
    \label{fig:appx_D2O_models2}
\end{figure}

\clearpage
\newpage
\section{Sample errors made by the resnext101\_32x8d\_wsl model}
\label{appx:errors}

Here we show sample errors made by resnext101\_32x8d\_wsl model over different categories. A image is considered an error if the ground-truth is not within the top 5 predictions.\newline

\begin{figure}[htbp]
\centering 
 \includegraphics[width=1\linewidth]{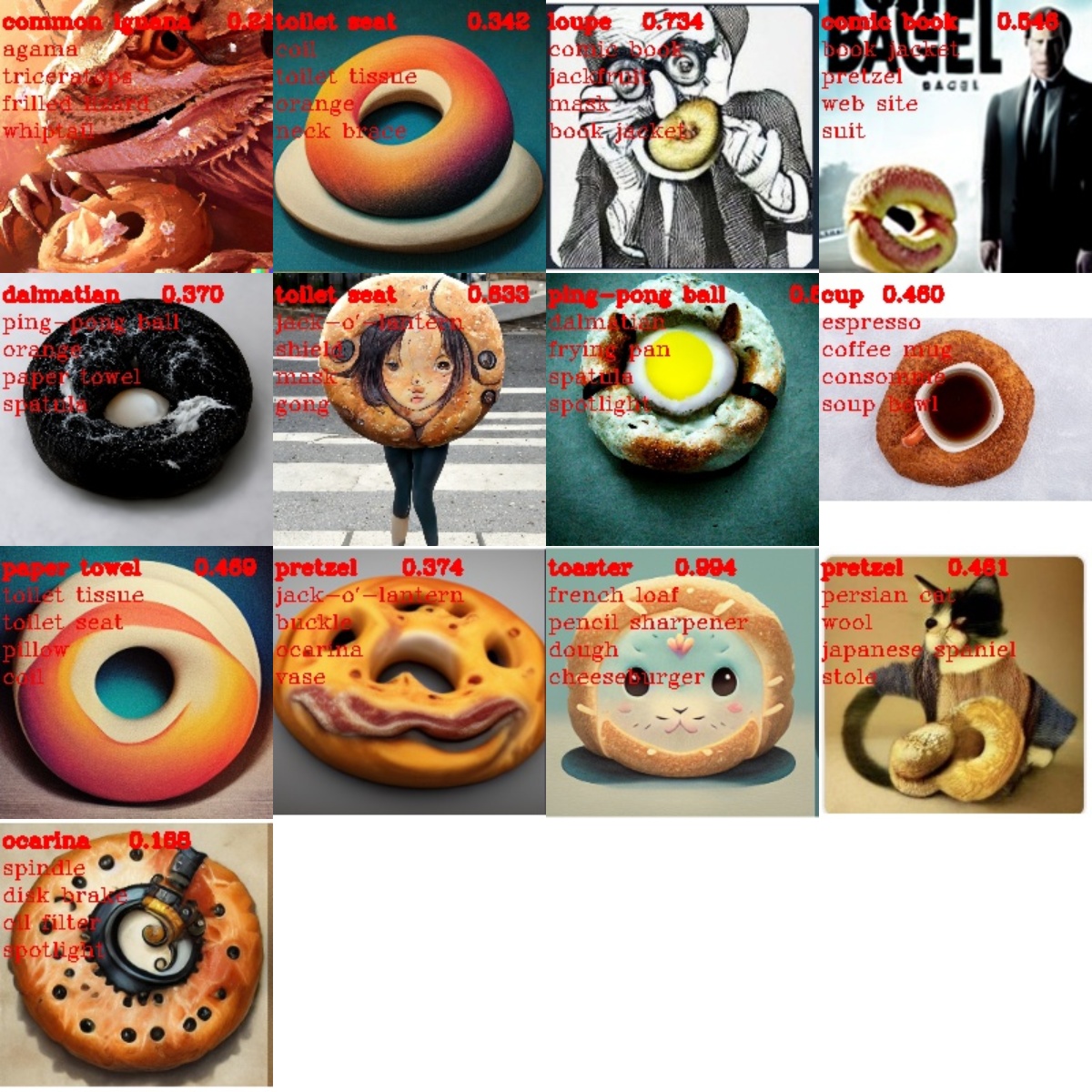}
\caption{Predictions of the resnext101\_32x8d\_wsl model over the bagel category.} 
\end{figure}

\begin{figure}[htbp]
\centering 
 \includegraphics[width=1\linewidth]{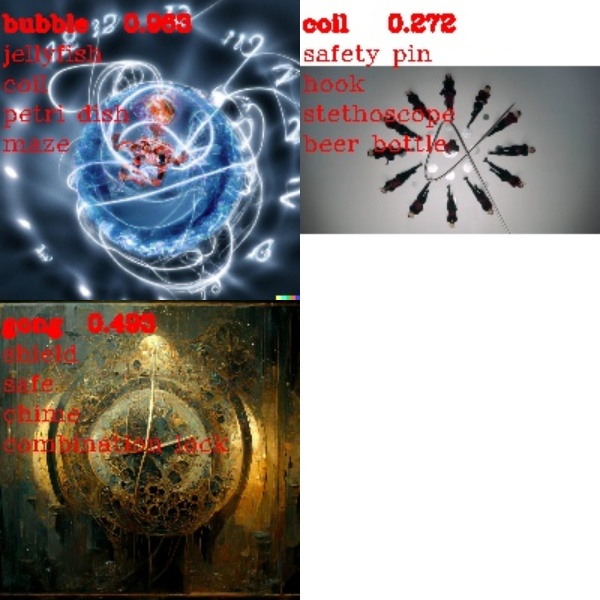}
\caption{Predictions of the resnext101\_32x8d\_wsl model over the clock category.} 
\end{figure}

\begin{figure}[htbp]
\centering 
 \includegraphics[width=1\linewidth]{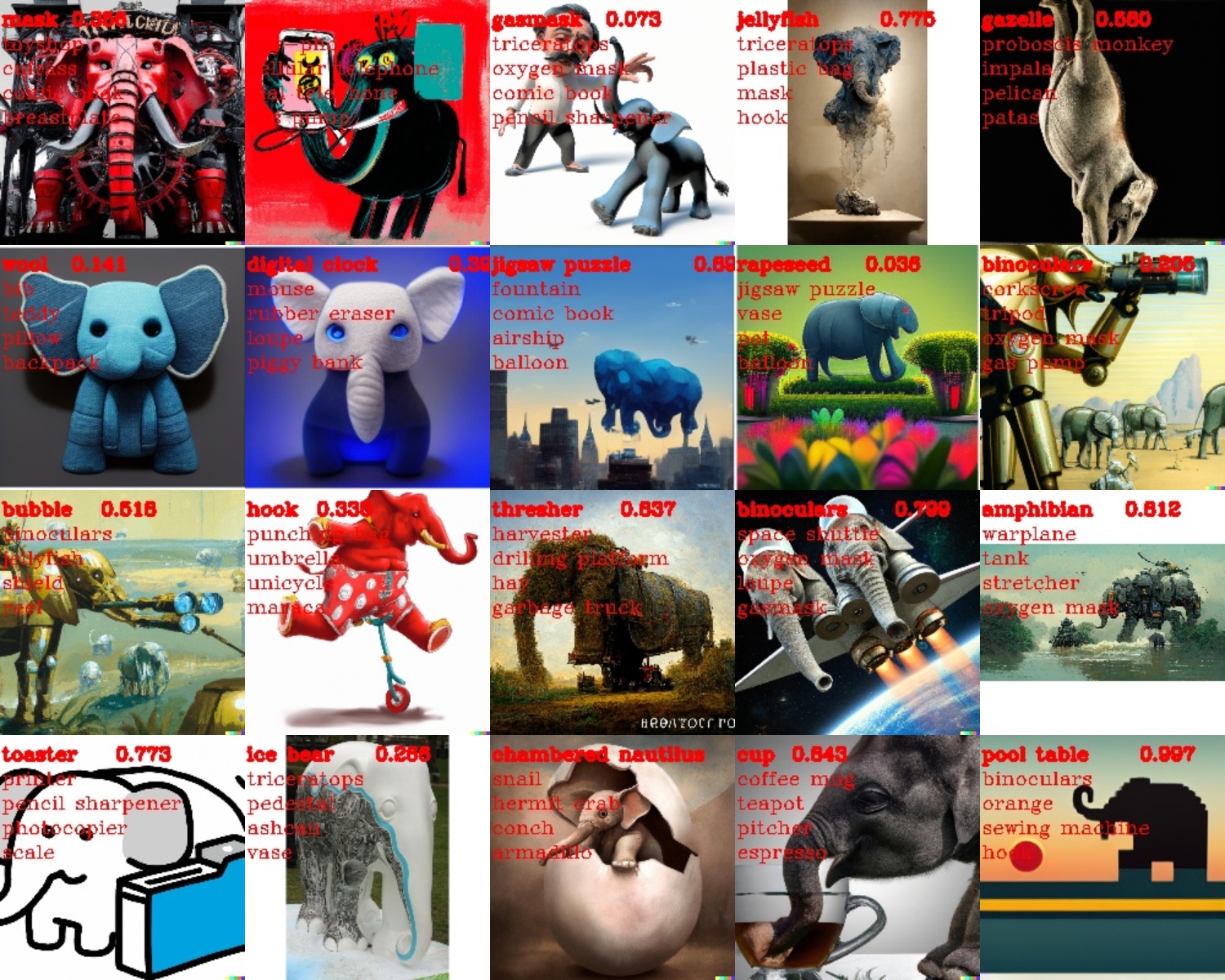}
\caption{Predictions of the resnext101\_32x8d\_wsl model over the elephant category.} 
\end{figure}

\begin{figure}[htbp]
\centering 
 \includegraphics[width=1\linewidth]{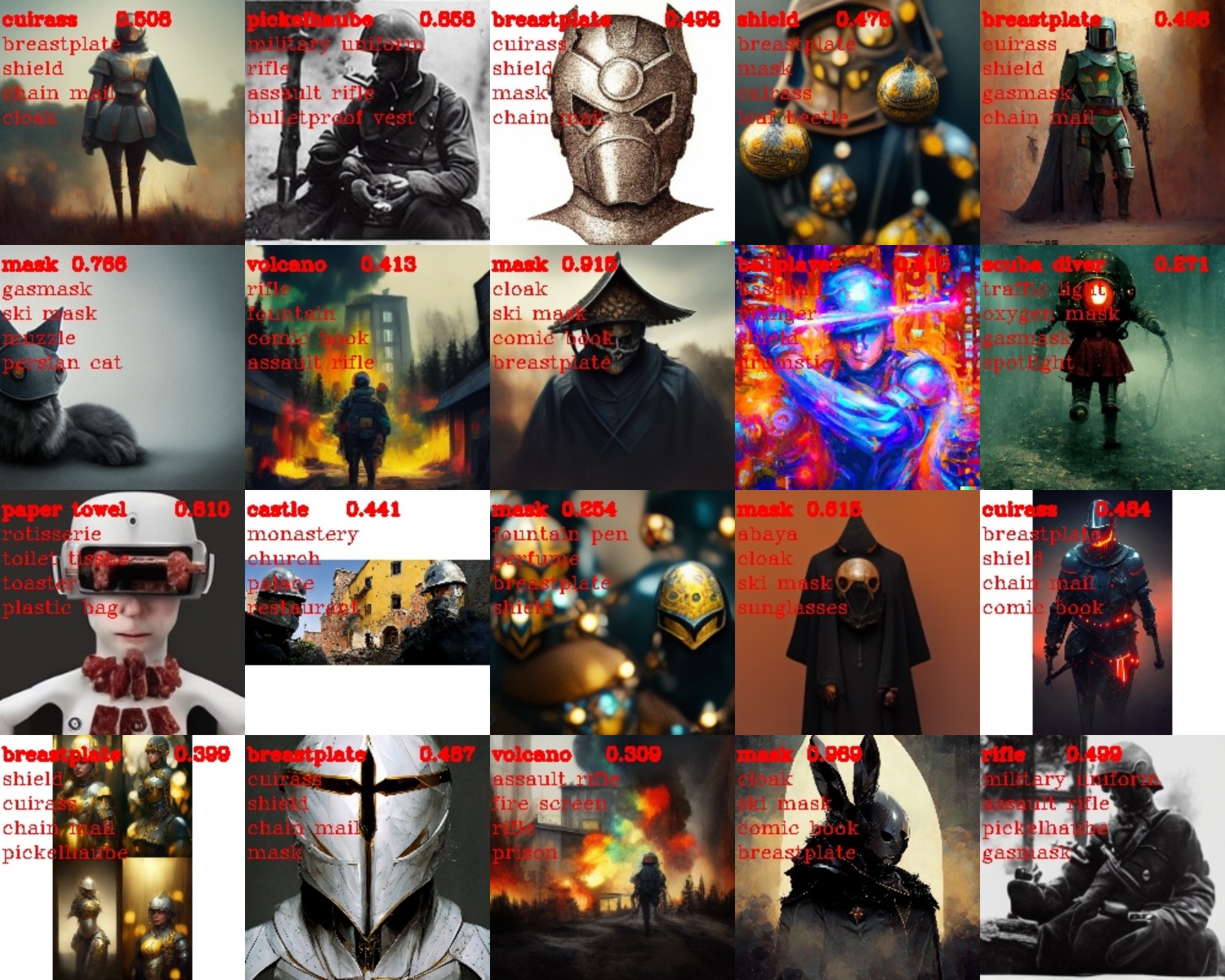}
\caption{Predictions of the resnext101\_32x8d\_wsl model over the helmet category.} 
\end{figure}

\begin{figure}[htbp]
\centering 
 \includegraphics[width=1\linewidth]{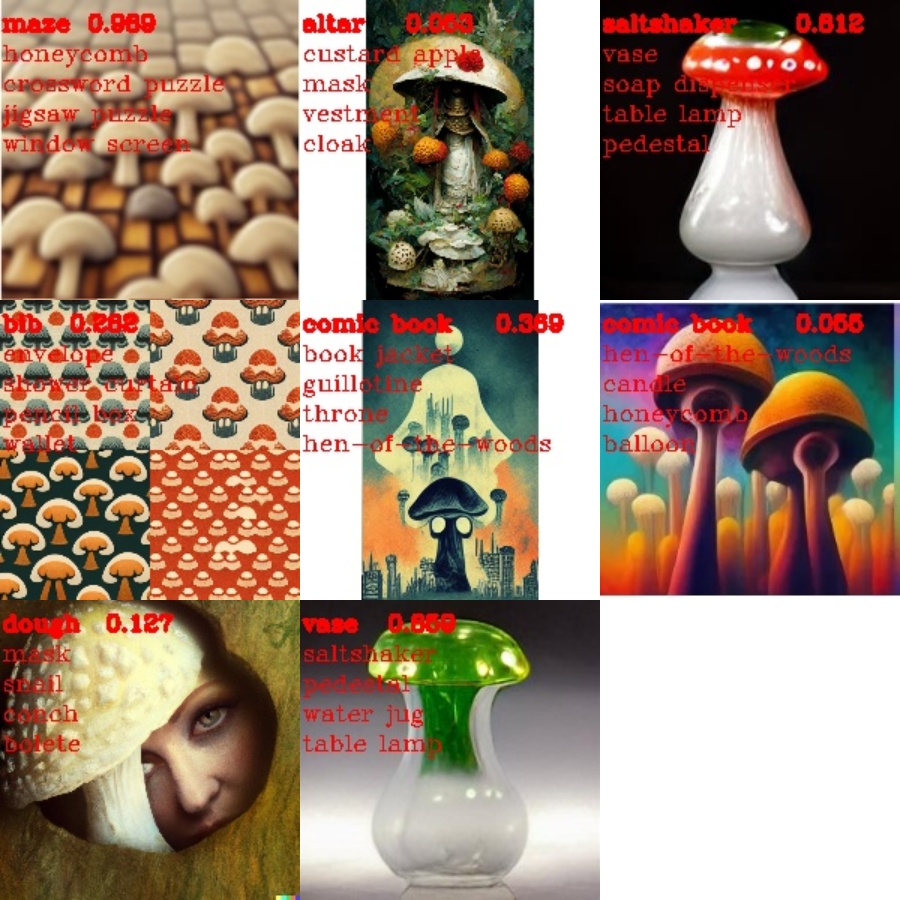}
\caption{Predictions of the resnext101\_32x8d\_wsl model over the mushroom category.} 
\end{figure} 

\begin{figure}[htbp]
\centering 
 \includegraphics[width=1\linewidth]{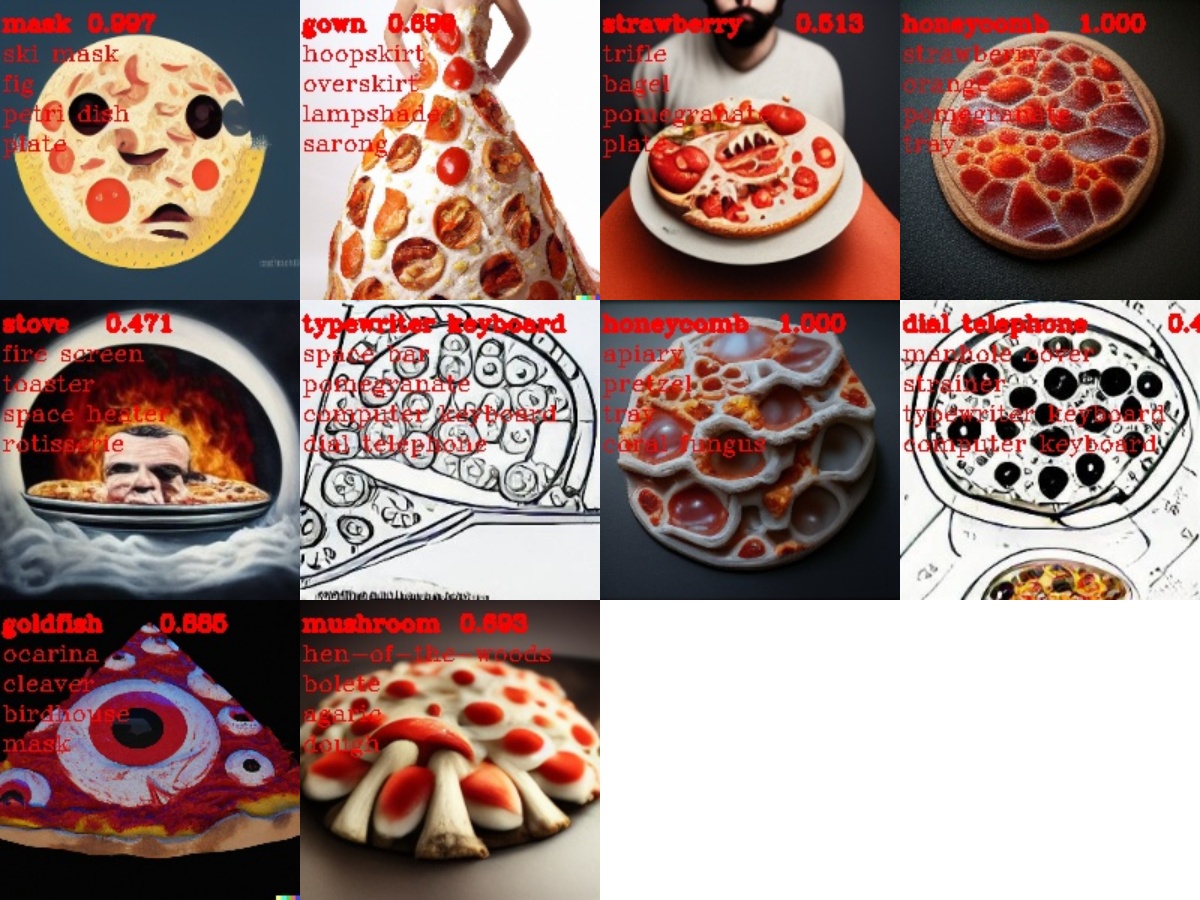}
\caption{Predictions of the resnext101\_32x8d\_wsl model over the pizza category.} 
\end{figure} 

\begin{figure}[htbp]
\centering 
 \includegraphics[width=1\linewidth]{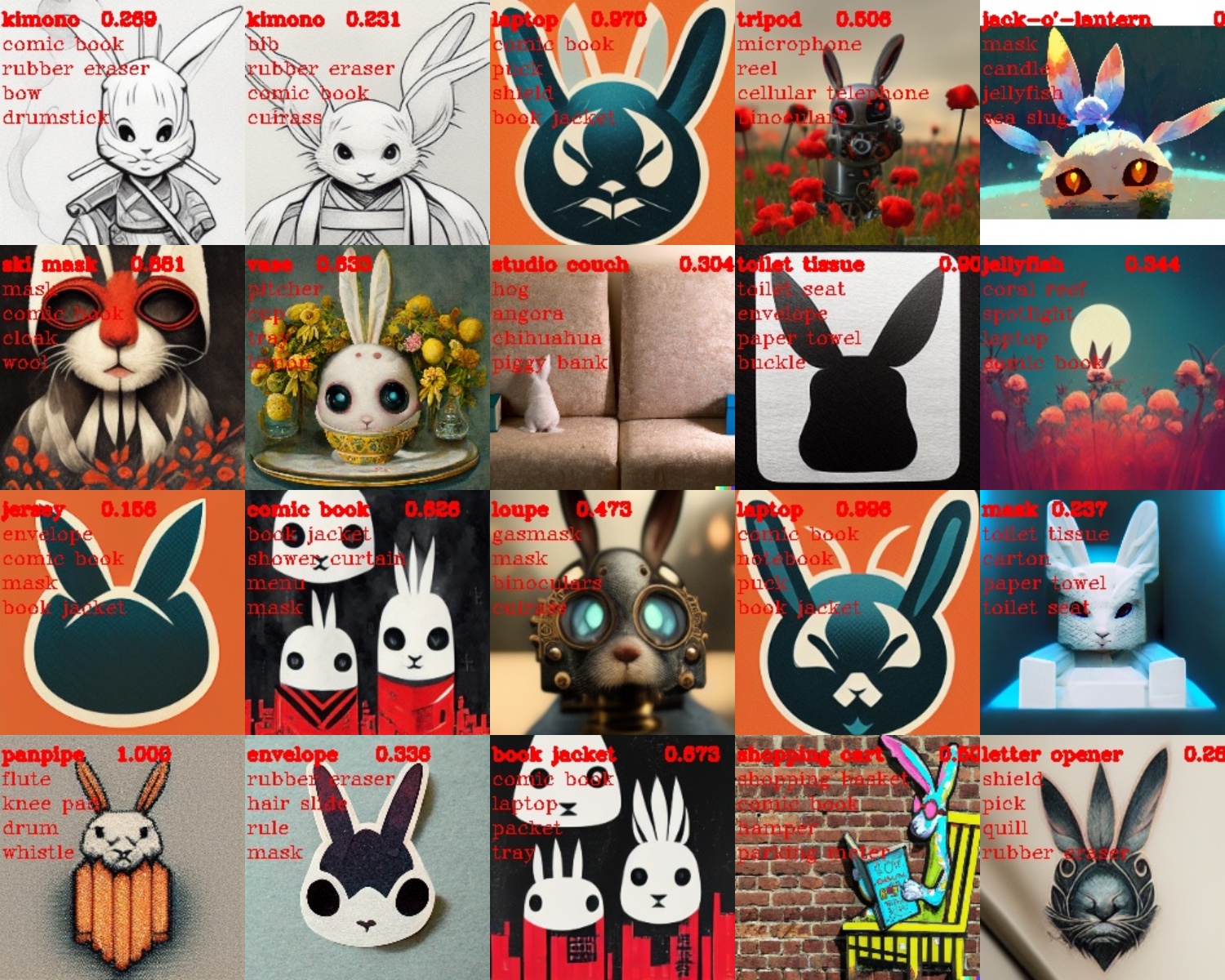}
\caption{Predictions of the resnext101\_32x8d\_wsl model over the rabbit/hare category.} 
\end{figure}

\begin{figure}[htbp]
\centering 
 \includegraphics[width=1\linewidth]{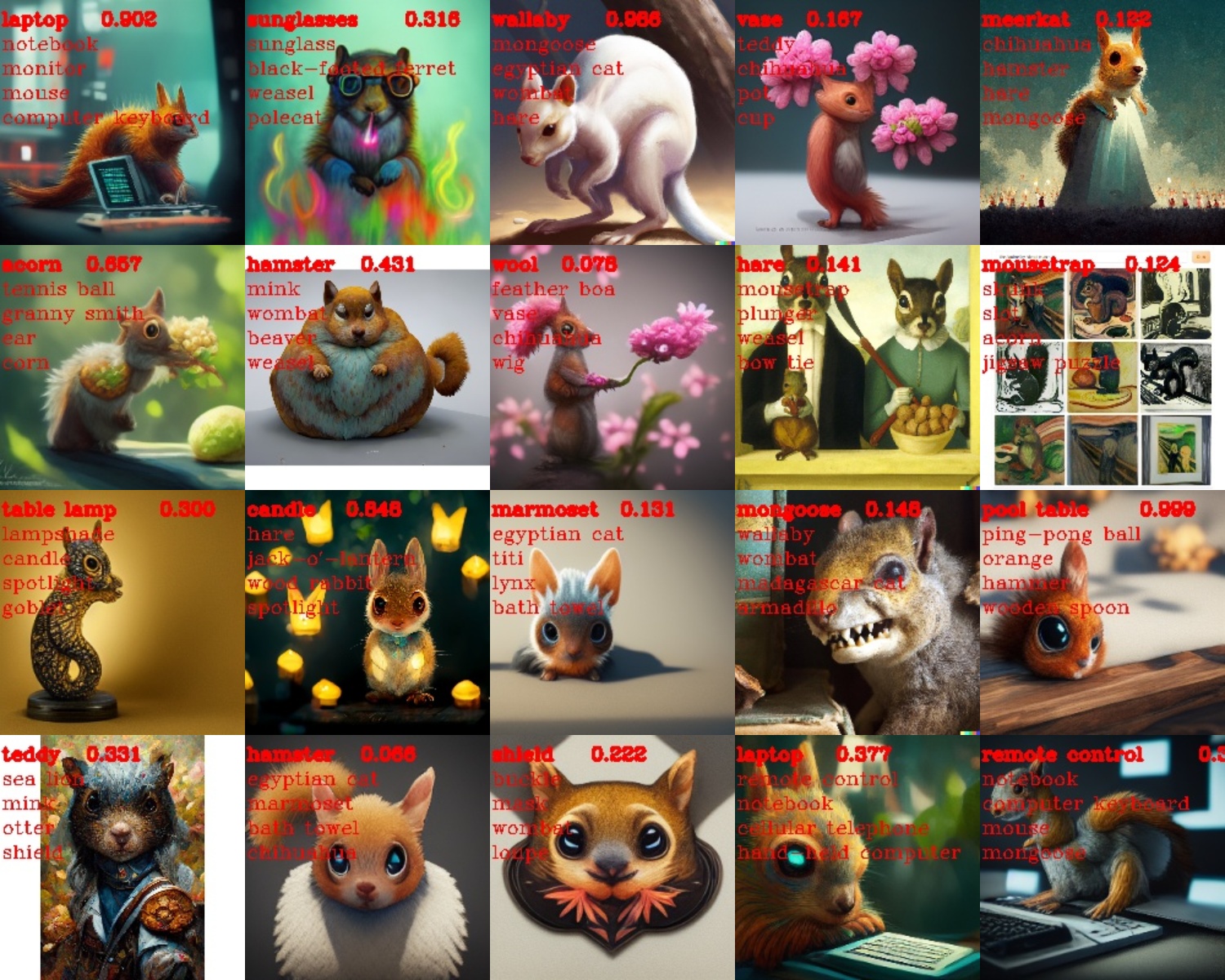}
\caption{Predictions of the resnext101\_32x8d\_wsl model over the squirrel category.} 
\end{figure}

\begin{figure}[htbp]
\centering 
 \includegraphics[width=1\linewidth]{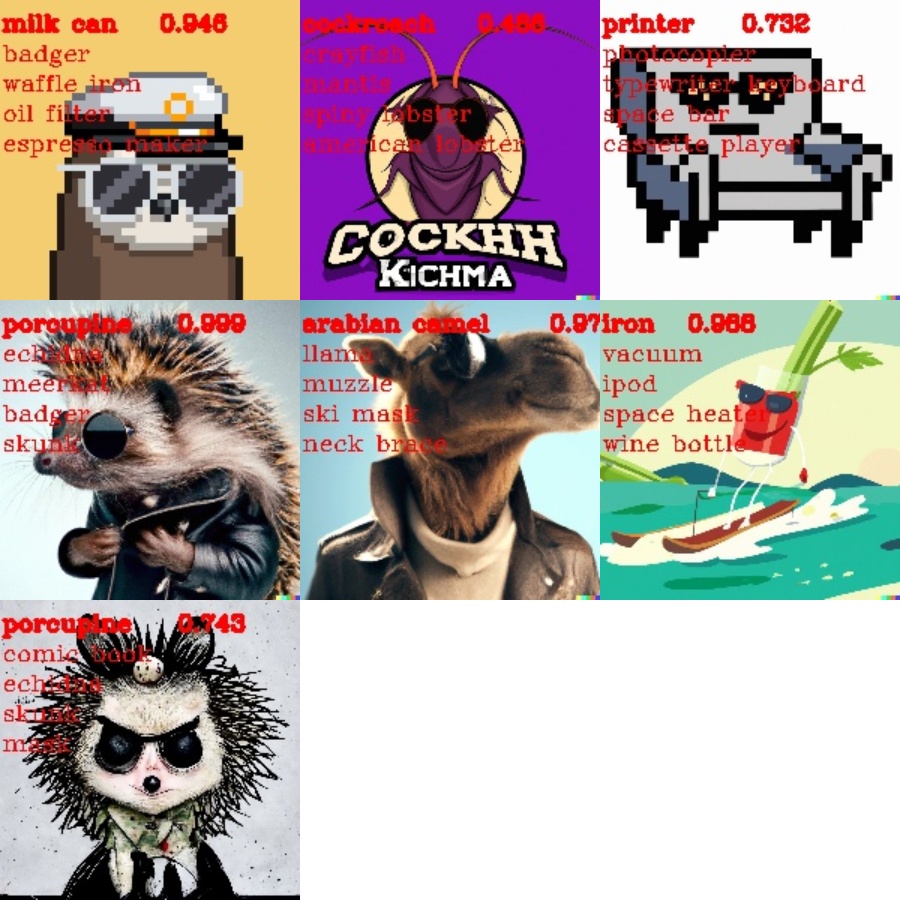}
\caption{Predictions of the resnext101\_32x8d\_wsl model over the sunglass category.} 
\end{figure}

\begin{figure}[htbp]
\centering 
 \includegraphics[width=.7\linewidth]{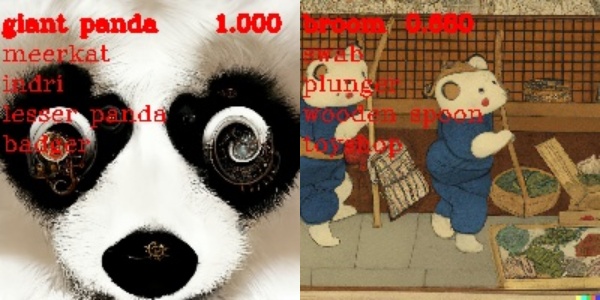}
\caption{Predictions of the resnext101\_32x8d\_wsl model over the teddy category.} 
\end{figure}

\begin{figure}[htbp]
\centering 
 \includegraphics[width=.33\linewidth]{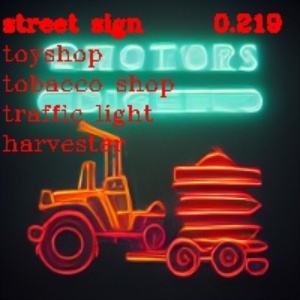}
\caption{Predictions of the resnext101\_32x8d\_wsl model over the tractor category.} 
\end{figure}

\begin{figure}[htbp]
\centering 
 \includegraphics[width=1\linewidth]{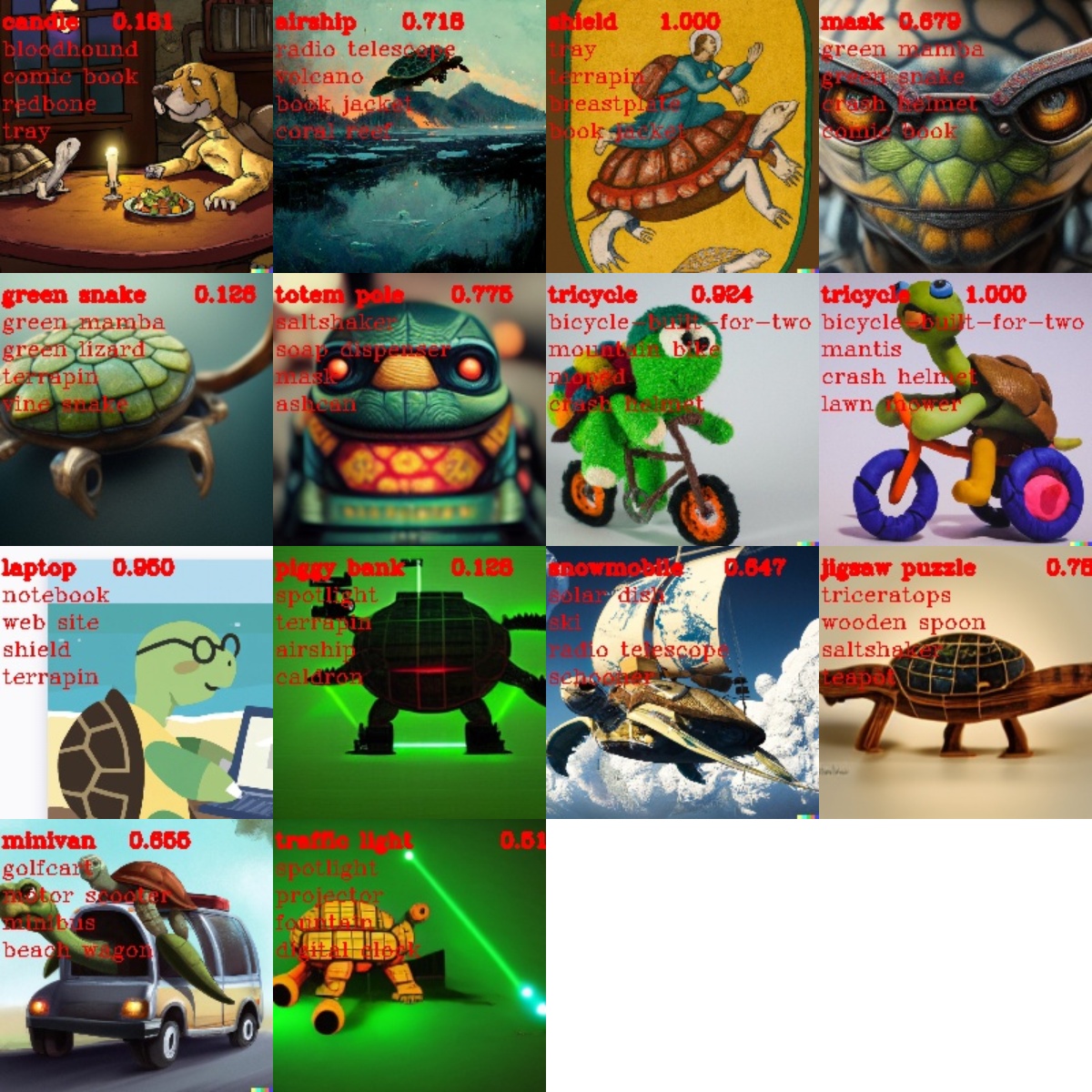}
\caption{Predictions of the resnext101\_32x8d\_wsl model over the turtle category.} 
\end{figure}

\begin{figure}[htbp]
\centering 
 \includegraphics[width=1\linewidth]{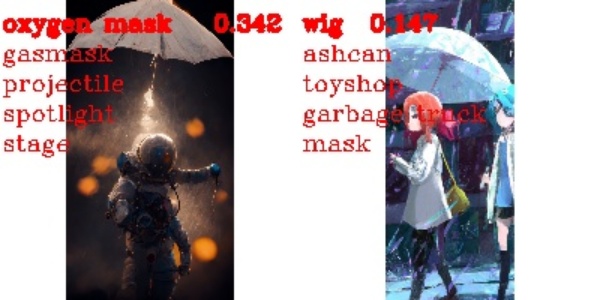}
\caption{Predictions of the resnext101\_32x8d\_wsl model over the umbrella category.} 
\end{figure}

\begin{figure}[htbp]
\centering 
 \includegraphics[width=1\linewidth]{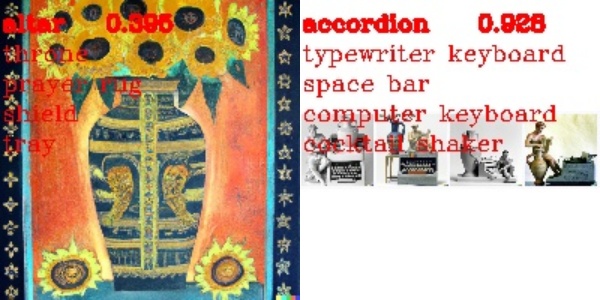}
\caption{Predictions of the resnext101\_32x8d\_wsl model over the vase category.} 
\end{figure}

\newpage
\section{Additional VQA results}
\label{appx:vqa}

\begin{figure}[htbp]
\centering 
 \includegraphics[width=1\linewidth]{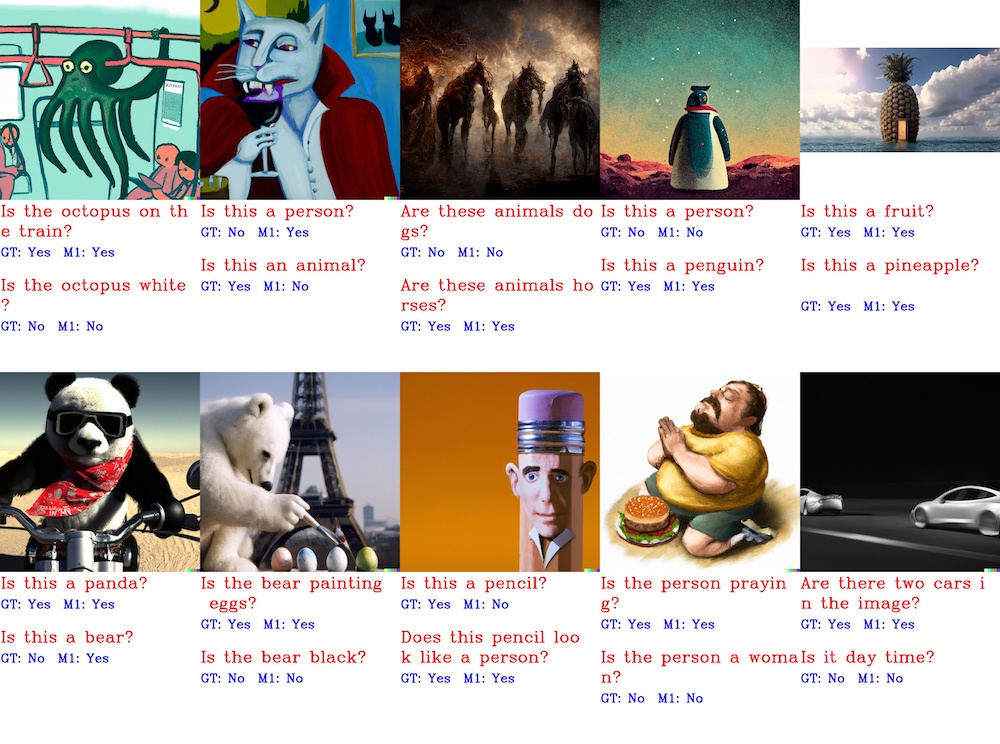}
 \includegraphics[width=1\linewidth]{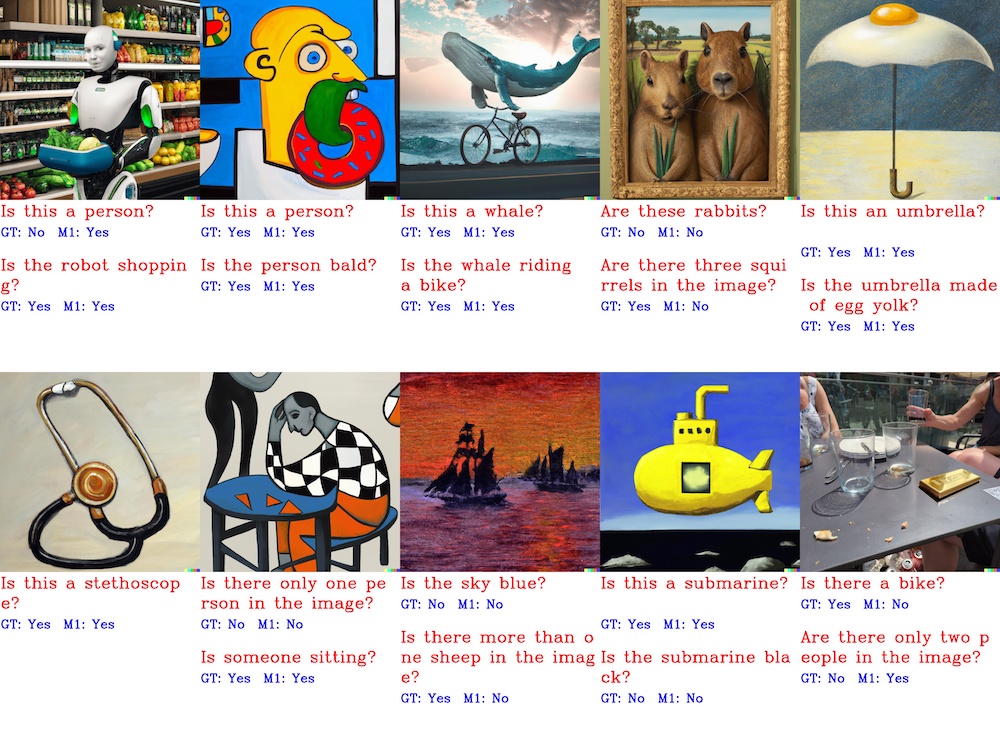}
\caption{Sample images and questions along with the predictions of the OFA model.} 
\label{fig:vqa1}
\vspace{-150pt}
\end{figure}

\begin{figure}[t]
\centering 
 \includegraphics[width=1\linewidth]{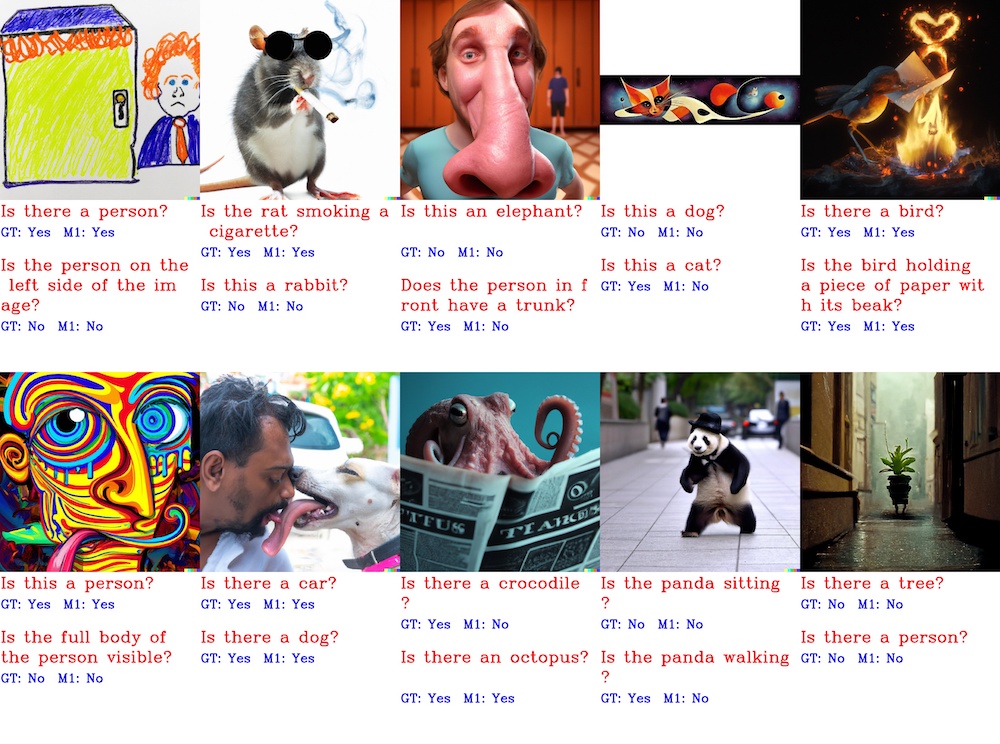}
 \includegraphics[width=1\linewidth]{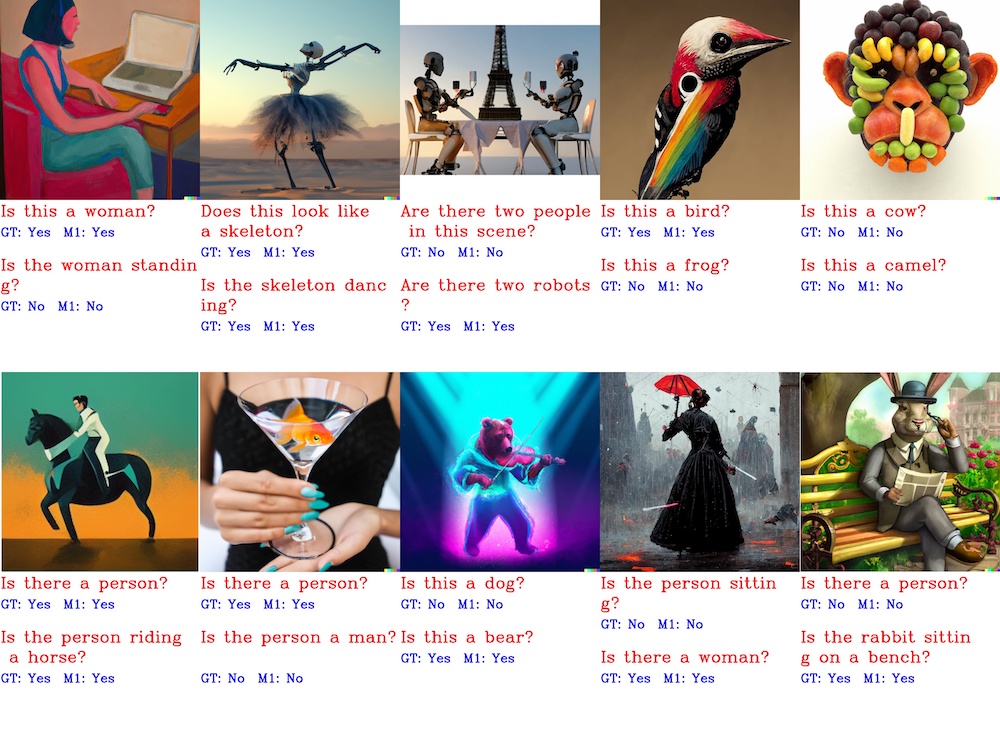}
\caption{Sample images and questions along with the predictions of the OFA model (cnt'd).} 
\label{fig:vqa2}
\end{figure}

\end{document}